\documentclass[letterpaper]{article} 
\usepackage{aaai2026}  
\usepackage{times}  
\usepackage{helvet}  
\usepackage{courier}  
\usepackage[hyphens]{url}  
\usepackage{graphicx} 
\usepackage{natbib}  
\usepackage{caption} 
\usepackage{algorithm}
\usepackage{algorithmic}

\usepackage{newfloat}
\usepackage{listings}
\DeclareCaptionStyle{ruled}{labelfont=normalfont,labelsep=colon,strut=off} 
\floatstyle{ruled}
\newfloat{listing}{tb}{lst}{}
\floatname{listing}{Listing}

\usepackage{subcaption}
\usepackage{makecell}

\usepackage{pifont}
\usepackage{amssymb}
\usepackage{booktabs}
\newcommand{\cmark}{{\ding{51}}}%
\newcommand{\xmark}{{\ding{55}}}%

\usepackage{multirow}

\newcommand{\acc}[2]{#1$_{\pm\text{#2}}$}
\usepackage{colortbl}
\usepackage{amsmath}
\usepackage[dvipsnames]{xcolor}
\definecolor{LightCyan}{HTML}{F0F0F0}
\newcommand{\colorLC}{\cellcolor{LightCyan}}

\newcommand{\dsa}{\cite{pmlr-v139-zhao21a}}
\newcommand{\dc}{\cite{zhao2021dataset}}
\newcommand{\mtt}{\cite{cazenavette2022distillation}}

\newcommand{\datm}{\cite{guo2024towards}}

\newcommand{\dm}{\cite{zhao2022dataset}}

\DeclareMathOperator*{\argminA}{arg\,min} 

\newcommand{\eg}{\textit{e.g.}}

\newcommand{\reffig}[1]{Figure \ref{#1}}
\newcommand{\reftab}[1]{Table \ref{#1}}

\title{Post Training Quantization for Efficient Dataset Condensation}
\author{
   Linh-Tam Tran, and Sung-Ho Bae\thanks{Corresponding Author}
}
\affiliations{
    Kyung Hee University, Republic of Korea\\

    \{tamlt,shbae\}@khu.ac.kr
}

\begin{document}

\maketitle

\begin{abstract}

Dataset Condensation (DC) distills knowledge from large datasets into smaller ones, accelerating training and reducing storage requirements. However, despite notable progress, prior methods have largely overlooked the potential of quantization for further reducing storage costs. In this paper, we take the first step to explore post-training quantization in dataset condensation, demonstrating its effectiveness in reducing storage size while maintaining representation quality without requiring expensive training cost. However, we find that at extremely low bit-widths (e.g., 2-bit), conventional quantization leads to substantial degradation in representation quality, negatively impacting the networks trained on these data.
To address this, we propose a novel \emph{patch-based post-training quantization} approach that ensures localized quantization with minimal loss of information. To reduce the overhead of quantization parameters, especially for small patch sizes, we employ quantization-aware clustering to identify similar patches and subsequently aggregate them for efficient quantization. Furthermore, we introduce a refinement module to align the distribution between original images and their dequantized counterparts, compensating for quantization errors.
Our method is a plug-and-play framework that can be applied to synthetic images generated by various DC methods. Extensive experiments across diverse benchmarks including CIFAR-10/100, Tiny ImageNet, and ImageNet subsets demonstrate that our method consistently outperforms prior works under the same storage constraints. Notably, our method nearly \textbf{doubles the test accuracy} of existing methods at extreme compression regimes (e.g., 26.0\% $\rightarrow$ 54.1\% for DM at IPC=1), while operating directly on 2-bit images without additional distillation. 
\end{abstract}

\section{Introduction}
Deep Neural Networks (DNNs) have become a primary solution in many computer vision tasks \cite{resnet,Redmon_2016_CVPR}. However, training these networks effectively requires vast amounts of data and significant computational resources \cite{vgg,resnet,Redmon_2016_CVPR}. To address this, Dataset Condensation (DC) \cite{DBLP:journals/corr/abs-1811-10959,zhao2021dataset} techniques have been developed to create smaller, condensed datasets that preserve the performance of the original, larger dataset. Due to its effectiveness, DC is an active research topic \cite{zhao2021dataset,cazenavette2022distillation,9879629,yuan2024colororiented} with numerous applications, including neural architecture search \dc, continual learning \cite{yang2023an,gu2024summarizingstreamdatamemoryconstrained}, super-resolution \cite{DBLP:conf/aaai/ZhangSZSZ24}, and privacy preserving \cite{pmlr-v162-dong22c,10622462}.

\begin{figure}[]
\centering

\includegraphics[height=1.3in]{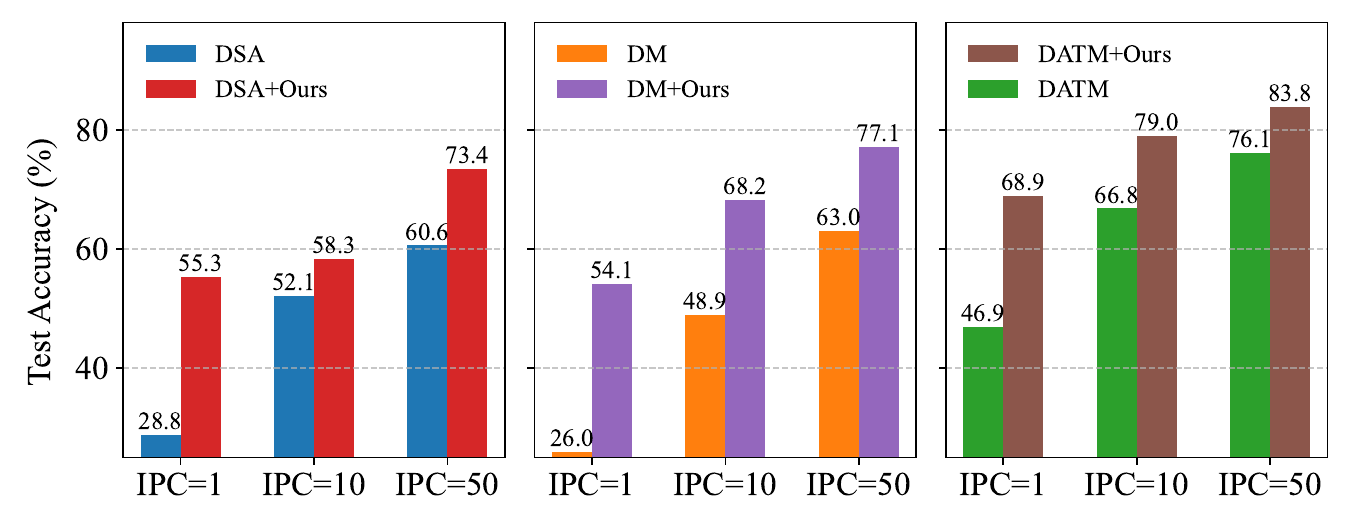}

\caption{Performance comparison when applying our framework to DSA \dsa, DM \dm, and DATM \datm. Notably, our framework \textbf{doubles the performance} for DSA and DM at a budget of 1 Image Per Class (IPC), demonstrating its effectiveness in extremely low-storage scenarios.}
\label{fig1}
\end{figure}

Approaches such as gradient matching \dc, distribution matching \dm, and trajectory matching \mtt\ have demonstrated strong performance in producing high-quality condensed datasets. However, these methods primarily focus on improving dataset quality while neglecting storage efficiency, as each synthetic sample still requires full-resolution storage. 

To overcome such storage bottlenecks, Parameterization-based Dataset Condensation (PDC) methods encode synthetic data as compact latent representations instead of raw pixels \cite{yuan2024colororiented}. These include frequency-space learning \cite{shin2023frequency}, sample factorization \cite{liu2022dataset}, and color space optimization \cite{yuan2024colororiented}. By reconstructing images at inference time, PDC methods achieve higher compression while retaining task-relevant information, often surpassing conventional DC in efficiency and scalability.

However, to the best of our knowledge, no prior work has explored post-training quantization (PTQ) to compress synthetic images for dataset condensation, despite its potential to significantly reduce storage costs with minimal overhead. This motivates our work, which aims to integrate PTQ into DC, reducing the bit-width of stored synthetic images without requiring expensive retraining or decoder networks.

To this end, we propose a novel framework that stores synthetic images in extremely low bit-widths (\eg, 2-bit). A naive application of global quantization—using a single set of quantization parameters for an entire image—suffers from severe degradation of task-relevant information due to spatially varying textures and details being quantized with insufficient precision. This results in a dramatic drop in the network’s performance when trained on the dequantized images.

To address this, we propose a \emph{patch-based asymmetric quantization} scheme that applies quantization locally to each image patch, effectively preserving spatial variations. To further reduce the overhead of quantization parameters, we introduce a quantization-aware grouping strategy that clusters similar patches and applies shared quantization within each group. Finally, we present a refinement module that finetunes the synthetic images to align their feature distributions with those of the original images, mitigating the effects of quantization noise.

By applying our method to various synthetic images generated from existing DC techniques, we observe up to \textbf{2$\times$ accuracy improvement at IPC=1} (see \reffig{fig1}) under extreme storage constraints, while operating entirely with 2-bit images.

Our contributions are summarized as follows:
\begin{itemize}
    \item We introduce a patch-based quantization method that significantly improves task performance at extremely low bit-widths by preserving spatial and structural details.
    \item We propose a clustering-based grouping strategy to share quantization parameters among similar patches, reducing storage overhead.
    \item We present a quantization-aware refinement module that aligns the distributions of dequantized and original synthetic images to improve downstream model performance.
    \item We validate our method through extensive experiments on diverse datasets, consistently achieving state-of-the-art performance under constrained storage budgets.
\end{itemize}

\section{Related Works}

\subsection{Dataset Condensation and Redundancy}

Dataset Condensation (DC) aims to distill a large dataset into a compact synthetic set such that a model trained on it performs comparably to one trained on the full dataset \cite{DBLP:journals/corr/abs-1811-10959,zhao2021dataset}. Early approaches address this by surrogate optimization, such as matching the gradients \cite{zhao2021dataset,pmlr-v162-kim22c,pmlr-v139-zhao21a}, network parameters \cite{cazenavette2022distillation,cui2023scaling,Du_2023_CVPR,guo2024towards}, or intermediate feature distributions \cite{zhao2022dataset,Zhao_2023_CVPR,9879629,datadam} between real and synthetic data.

While these methods successfully preserve task performance, they are inefficient in storage since the synthesized images are stored at full resolution and precision. To overcome this, PDC explores compact representations by reducing various forms of redundancy:

\begin{itemize}
    \item \textbf{IDC} reduces \textit{spatial redundancy} by storing data in low-dimensional latent spaces and reconstructing them via upsampling \cite{pmlr-v162-kim22c}.
    \item \textbf{AutoPalette} targets \textit{color redundancy} using a learned palette encoder \cite{yuan2024colororiented}.
    \item \textbf{DDiF} compresses \textit{feature redundancy} by encoding data into neural fields that generate images on demand \cite{ddif}.
\end{itemize}

Although these methods improve storage efficiency, they still rely on $32$-bit representations and incur high computational cost due to decoder networks. In contrast, our method uniquely targets \textit{bit-level redundancy} through PTQ, enabling extremely low-bit (\eg, $2$-bit) storage. Table~\ref{tab:redundancy_comparison} summarizes recent PDC methods in terms of their redundancy type, compression strategy, approximate compression ratio (vs. $32$-bit baseline), storage bitwidth, and computational overhead. Our method achieves the highest ratio via bit-level quantization while maintaining the lowest overhead.

\begin{table}[t]
\centering
\footnotesize
\setlength{\tabcolsep}{4pt}
\begin{tabular}{l|l|l|c|c}
\toprule
\textbf{Method} & \textbf{Redundancy} & \textbf{Strategy}& \textbf{Bit} & \textbf{Cost} \\
\midrule
IDC & Spatial & Spatial Downsample & $32$ & Med. \\
AutoPalette & Color & Palette Reduction & $32$ & High \\
DDiF & Feature & Neural Field & $32$ & High \\
\textbf{Ours} & Bit-level & Patch Quantization & $\textbf{2}$ & \textbf{Low} \\
\bottomrule
\end{tabular}
\caption{Comparison of redundancy reduction strategies in PDC.}
\label{tab:redundancy_comparison}
\end{table}

\subsection{Post-Training Quantization}

Quantization reduces storage and computation by representing model weights or activations with fewer bits \cite{han2015deep_compression,Esser2020LEARNED}. PTQ, in particular, is a widely adopted technique for efficient deployment on resource-constrained devices, since it avoids retraining \cite{DBLP:journals/corr/abs-2201-11113,PTQ4ViT_arixv2022,cai2020zeroq,yao2023zeroquantv2exploringposttrainingquantization}.

While PTQ has been extensively explored for compressing models, its application to dataset condensation has remained largely unexplored. To the best of our knowledge, this is the first work that applies post-training quantization to synthetic datasets in dataset condensation. Our method introduces a patch-based quantization framework with grouping and refinement, operating directly on 2-bit tensors without additional training or decoder models. This enables highly compressed, efficient, and distillation-free dataset storage and deployment.

\begin{figure*}[t]
\centering
\includegraphics[width=0.95\linewidth]{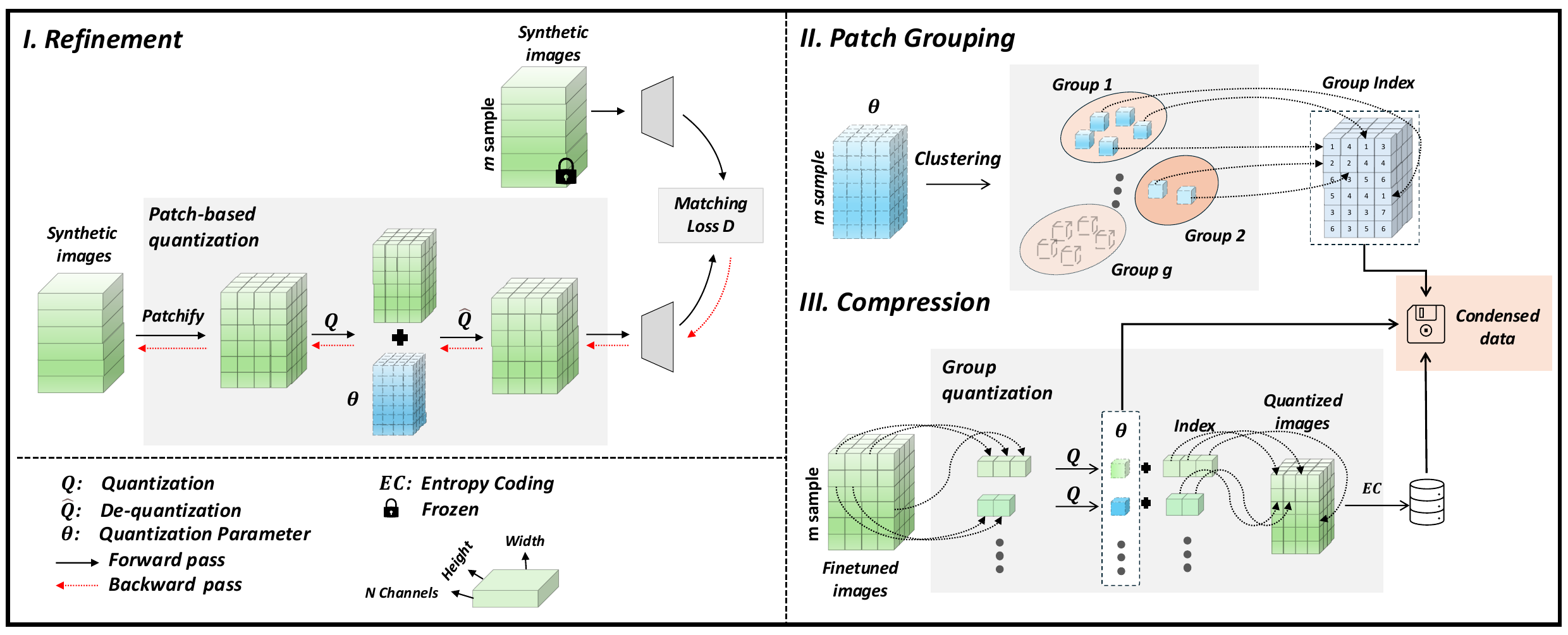}
\caption{
Overview of the proposed patch-based quantization framework for dataset condensation. 
(I) Synthetic images are first refined using a quantization-aware loss. 
(II) Patches are grouped based on quantization parameters via $k$-means clustering. 
(III) Each group is quantized with shared parameters, and the result is entropy encoded to produce the final compressed dataset.
}
\label{overview}
\end{figure*}

\section{Method}
\reffig{overview} provides the overview of our proposed framework. In this section, we first provide background on DC and quantization. Then, we introduce our patch-based quantization. Next, we present our quantization grouping. Lastly, we describe the refinement module.

\subsection{Preliminaries}
\paragraph{Dataset Condensation.} DC aims to generate a compact synthetic dataset $\mathcal{S} = \{(x^i, y^i) \mid i = 0, 1, \dots, |\mathcal{S}| - 1\}$ from a larger dataset $\mathcal{T} = \{(x^i, y^i) \mid i = 0, 1, \dots, |\mathcal{T}| - 1\}$, such that training a model on $\mathcal{S}$ yields performance comparable to training on $\mathcal{T}$. Let $\mathcal{D}$ denote a distance metric (e.g., Mean Squared Error), and $\Phi$ denote a matching objective function. The condensation problem is formulated as the following optimization:
\begin{equation}
\mathcal{S}^* = \argminA_{\mathcal{S}} \mathcal{D} \left( \Phi(\mathcal{S}), \Phi(\mathcal{T}) \right),
\label{generalformofdc}
\end{equation}
where $\mathcal{S}^*$ is the optimal condensed set minimizing the discrepancy between the two objective functions.

\paragraph{Quantization.} Quantization reduces the precision of model parameters by representing them with lower bit-width integers. A widely used method is \textit{asymmetric quantization} (AQ)~\cite{DBLP:journals/corr/abs-2106-08295}, which employs two parameters: the \textit{scale} and the \textit{zero-point}. These enable the mapping of floating-point values to discrete integers. Let $\mathbf{x} \in \mathbb{R}^{H \times W \times C}$ be an input tensor, and let $Q_{\text{min}} = 0$ and $Q_{\text{max}} = 2^{b} - 1$ denote the quantization bounds for a bit-width $b$. The scale factor $\alpha$ is computed as:
\begin{equation}
    \alpha = \frac{\max(\mathbf{x}) - \min(\mathbf{x})}{Q_{\text{max}} - Q_{\text{min}}}
\end{equation}

The zero-point $z$ aligns the minimum value to the quantized range:
\begin{equation}
z = \left\lfloor Q_{\text{min}} - \frac{\min(\mathbf{x})}{\alpha} \right\rceil
\end{equation}
where $\left\lfloor \cdot \right\rceil$ denotes rounding to the nearest integer. Given $\alpha$ and $z$, quantization and dequantization are defined as:
\begin{equation}
\mathbf{x}^q = \left\lfloor \frac{\mathbf{x}}{\alpha} + z \right\rceil, \quad \mathbf{x}^{\text{deq}} = (\mathbf{x}^q - z) \cdot \alpha.
\end{equation}

\begin{figure*}[t]
\centering

\includegraphics[width=0.95\textwidth]{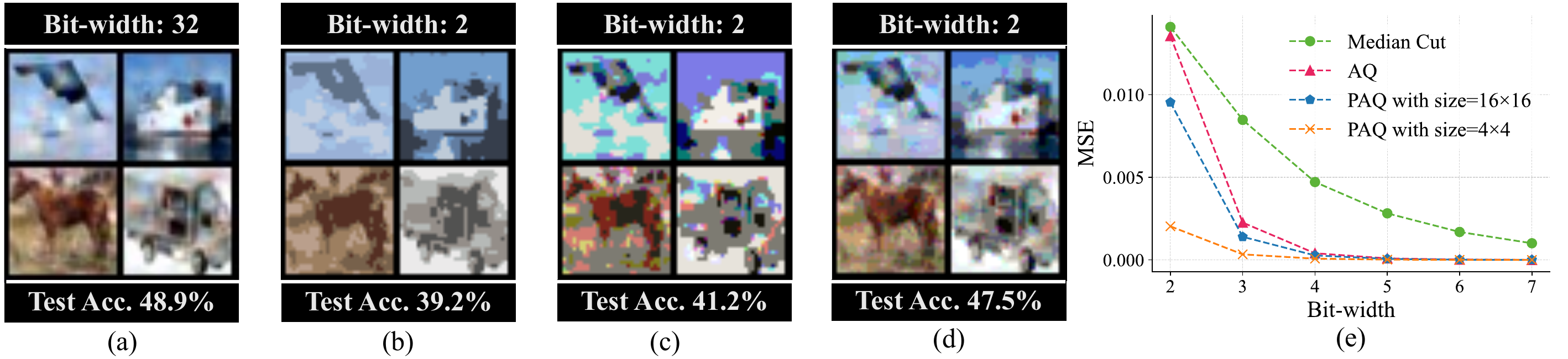}
\caption{
Visualization comparison of quantization strategies on synthetic images: 
(a) original images, (b) Median Cut~\citep{mediancut}, (c) asymmetric quantization, and (d) our patch-based quantization. 
All quantized images use 2-bit precision. (e) shows distortion (MSE) versus bit-width across methods. 
Images are generated by DM~\dm\ on CIFAR-10.
}
\label{challenge}
\end{figure*}

\subsection{Analysis of Whole-Image Quantization}
Whole-image quantization applies a single set of parameters (scale and zero-point) to the entire image. While simple and storage-efficient, this global approach suffers at extremely low bit-widths due to its inability to preserve fine-grained features. As a result, downstream performance degrades significantly.

We analyze this issue using two methods: (1) asymmetric quantization, and (2) Median Cut~\citep{mediancut}, which partitions color space recursively along high-variance channels. We apply each method to synthetic images generated by DM~\dm, then dequantize and train a 3-layer CNN for evaluation.

As shown in Fig.~\ref{challenge} (b,c), both whole-image quantization methods result in noticeable visual distortion and accuracy degradation compared to the original images (a). To quantify this, we measure the Mean Squared Error (MSE) between the feature embeddings of original and dequantized images. The results in Fig.~\ref{challenge} (e) confirm that representational distortion increases as bit-width decreases. These findings suggest that whole-image quantization fails to maintain essential information needed for learning under aggressive compression.

\subsection{Patch-based Quantization}
To address the limitations of whole-image quantization and better align with the goals of dataset condensation, we introduce \textbf{patch-based asymmetric quantization} (PAQ). Unlike traditional approaches that apply a single global quantization to the entire image, PAQ operates at the patch level, enabling localized control of quantization and better preservation of fine-grained details.

Formally, given an image \( \mathbf{x} \), we divide it into \( P \) non-overlapping patches \( \{ \mathbf{x}_i \}_{i=1}^{P} \), where each \( \mathbf{x}_i \in \mathbb{R}^{h \times w \times C} \) represents a spatial region. We then perform per-patch quantization:
\begin{equation}
\mathbf{x}_i^{q} = Q(\mathbf{x}_i, \theta_i),
\end{equation}
where \( Q(\cdot, \theta_i) \) is the quantization function and \( \theta_i = (\alpha_i, z_i) \) includes the scale and zero-point for patch \( i \). This localized formulation allows each patch to adapt to its own distribution, minimizing information loss.

As shown in Fig.~\ref{challenge} (e), PAQ significantly reduces quantization-induced distortion across bit-widths compared to whole-image quantization. Notably, even at 2-bit precision, PAQ achieves competitive accuracy (47.5\%) relative to full-precision data (48.9\%), demonstrating its effectiveness in preserving task-relevant features under extreme compression.

\subsection{Quantization-Aware Patch Grouping}
While PAQ improves fidelity by assigning independent quantization parameters to each patch, it also increases storage cost due to the need to store \( \theta_i \) for all patches. To mitigate this, we propose \textbf{group-aware quantization} (GAQ), which clusters similar patches to share quantization parameters.

We apply \( k \)-means clustering in the quantization parameter space to group patches with similar quantization behavior. Specifically, each patch \( \mathbf{x}_i \) has an associated quantization parameter \( \theta_i = (\alpha_i, z_i) \), where \( \alpha_i \) and \( z_i \) denote the scale and zero-point, respectively. Let \( \mathcal{C}_g \) denote the set of patches assigned to group \( g \), and \( G \) be the total number of groups. Our objective is to minimize the intra-group variance of quantization parameters, encouraging patches within each group to share a common parameter:

\begin{align}
\{\mathcal{C}_g^*, \theta_g^*\}_{g=1}^G 
&= \underset{\{\mathcal{C}_g\}, \{\theta_g\}}{\arg\min} 
\sum_{g=1}^{G} \sum_{\theta_i \in \mathcal{C}_g} 
\left\| \theta_i - \hat{\theta}_g \right\|^2, \\
&\text{where } \hat{\theta}_g = \frac{1}{|\mathcal{C}_g|} 
\sum_{\theta_i \in \mathcal{C}_g} \theta_i. \nonumber
\end{align}
where \( \hat{\theta}_g \) represents the centroid (mean) of the quantization parameters in group \( g \). This clustering step enables all patches in the same group to share a single quantization parameter \( \theta_g \), significantly reducing storage overhead while maintaining quantization fidelity.

\paragraph{Intra-group Recalibration.}
Rather than directly using the cluster centroids \( \hat{\theta}_g \) as quantization parameters, we recalibrate \( \theta_g \) based on the actual data distribution of patches within each group \( \mathcal{C}_g \). Specifically, for each group \( g \), we first concatenate all patches, \( \mathbf{x}_g = \text{concat}(\{ \mathbf{x}_i \}_{i \in \mathcal{C}_g}) \), and then reshape the result into a one-dimensional tensor, \( \mathbf{x}_g^{\text{flat}} = \text{reshape}(\mathbf{x}_g) \). We compute the group-specific quantization parameter \( \theta_g \) by calibrating over \( \mathbf{x}_g^{\text{flat}} \), and quantize the group as:
\begin{equation}
\mathbf{x}_g^q = Q(\mathbf{x}_g^{\text{flat}}, \theta_g),
\end{equation}
where \( Q(\cdot, \theta_g) \) is the quantization function. All patches in \( \mathcal{C}_g \) are quantized using this shared parameter \( \theta_g \). This recalibration ensures that the shared quantization parameters accurately reflect the underlying patch distributions, thereby reducing quantization error and enhancing downstream model performance.

\begin{table*}[]
\centering
\footnotesize
\begin{tabular}{l|l|ccc|ccc|cc}
\cmidrule[1pt]{1-10}
\multicolumn{2}{c|}{Dataset}& \multicolumn{3}{c|}{CIFAR-10} & \multicolumn{3}{c|}{CIFAR-100} &\multicolumn{2}{c}{Tiny ImageNet}  \\\cmidrule[0.5pt]{1-10}
\multicolumn{2}{c|}{IPC}  & 1 &10 &  50& 1&10&50 & 1 & 10\\\cmidrule[0.5pt]{1-10}

\multirow{5}{*}{Distillation} & DD & --& \acc{36.8}{1.2} &--&--&--&-- & -- & --\\
& DM &\acc{26.0}{0.8} &  \acc{48.9}{0.6} &\acc{63.0}{0.4}& \acc{11.4}{0.3} &\acc{29.7}{0.3}&\acc{43.6}{0.4} &\acc{3.9}{0.2} &\acc{12.9}{0.4}\\

& DSA & \acc{28.8}{0.7} &\acc{52.1}{0.5}&\acc{60.6}{0.5} & \acc{13.9}{0.3}&\acc{32.3}{0.3}&\acc{42.8}{0.4}  & -- & --\\
& MTT & \acc{46.3}{0.8}&\acc{65.3}{0.7}&\acc{71.6}{0.2}&\acc{24.3}{0.3}&\acc{40.1}{0.4}&\acc{47.7}{0.2} & \acc{8.8}{0.3} & \acc{23.2}{0.2} \\
& DATM  &\acc{46.9}{0.5}& \acc{66.8}{0.2}&\acc{76.1}{0.3}&\acc{27.9}{0.2}&\acc{47.2}{0.4}&\acc{55.0}{0.2} & \acc{17.1}{0.3} & \acc{31.1}{0.3}\\\cmidrule[0.5pt]{1-10}

\multirow{6}{*}{Parameterization} & IDC & \acc{55.0}{0.4}&\acc{67.5}{0.5}&\acc{74.5}{0.1}&--&--&-- &--&--\\
& HaBa  & \acc{48.3}{0.8} &\acc{69.9}{0.4}&\acc{74.0}{0.2}&\acc{33.4}{0.4}&\acc{40.2}{0.2}&\acc{47.0}{0.2} &-- & -- \\
& SPEED  & \acc{63.2}{0.1}&\acc{73.5}{0.2} &\acc{77.7}{0.4} &\acc{40.0}{0.4}&\acc{45.9}{0.3}&\acc{49.1}{0.2} & \acc{26.9}{0.3} & \acc{28.8}{0.2}\\
& FreD &\acc{60.6}{0.8} & \acc{70.3}{0.3} &\acc{75.8}{0.1} & \acc{34.6}{0.4} & \acc{42.7}{0.2}&\acc{47.8}{0.1}& \acc{19.2}{0.4} & \acc{24.2}{0.4} \\
&Spectral& \acc{68.5}{0.8}&\acc{73.4}{0.2} & \acc{75.2}{0.6}&\acc{36.5}{0.3}&\acc{46.1}{0.2} &-- & \acc{21.3}{0.2} & --\\
& AutoPalette & \acc{58.6}{1.1}&\acc{74.3}{0.2}&\acc{79.4}{0.2}&\acc{38.0}{0.1}&\acc{52.6}{0.3}&\acc{53.3}{0.8} & -- & --\\\cmidrule[0.5pt]{1-10}
\multirow{3}{*}{Quantization} 
 & DSA+Ours& \acc{55.3}{0.4} &\acc{58.3}{0.5}&\acc{73.4}{0.4}   &\acc{34.7}{0.3}& \acc{41.1}{0.5} &-- & -- & --\\
 & DM+Ours & \acc{54.1}{0.5} &\acc{68.2}{0.4} & \acc{77.1}{0.3}  &\acc{34.0}{0.3} &\acc{51.2}{0.3}&-- & \acc{14.8}{0.5} & \acc{30.6}{0.3}\\

& DATM+Ours & \colorLC\textbf{\acc{68.9}{0.4}}& \colorLC\textbf{\acc{79.0}{0.3}} & \colorLC\textbf{\acc{83.8}{0.2}}&\colorLC\textbf{\acc{48.0}{0.3}}&\colorLC\textbf{\acc{56.5}{0.2}}& -- & \colorLC\textbf{\acc{27.3}{0.5}} & \colorLC\textbf{\acc{39.4}{0.4}}\\\cmidrule[0.5pt]{1-10}

 & Whole $\mathcal{T}$ &\multicolumn{3}{c|}{\acc{84.8}{0.1}}  & \multicolumn{3}{c|}{\acc{56.2}{0.2}} & \multicolumn{2}{c}{\acc{37.6}{0.4}}
 \\
\cmidrule[1.pt]{1-10}
\end{tabular}
\caption{
Test accuracy (\%) comparison on CIFAR-10, CIFAR-100, and Tiny ImageNet. Results are reported in \textit{\acc{mean}{std}} format. Experiments on CIFAR-100 with IPC=50 are omitted, as this would require 1,000 images per class under a compression rate of 20, which exceeds the original 500 images per class.
}
\label{cifarexp}
\end{table*}

\subsection{Refinement for Synthetic Images} 
Our goal is to refine synthetic images to mitigate the feature-space distortion caused by quantization. Prior studies have shown that quantization noise in inputs can significantly degrade the quality of extracted features~\cite{8578928,Zheng_2016_CVPR,DBLP:journals/corr/VasiljevicCS16}. 

Given a target bit-width \( b \), we optimize a finetuned image \( \mathbf{x}^\text{ft} \) such that its feature representation remains close to that of the original synthetic image, even after undergoing quantization and dequantization. Formally, we define:
\begin{equation}
(\mathbf{x}^\text{ft})^\text{deq} = \hat{Q}(Q(\mathbf{x}^\text{ft}, \theta), \theta),
\end{equation}
where \( Q(\cdot, \theta) \) and \( \hat{Q}(\cdot, \theta) \) denote the quantization and dequantization functions parameterized by \( \theta \), respectively. 

We extract features from both the original and the dequantized versions using a neural network \( f(\cdot) \):
\begin{equation}
\mathbf{f} = f(\mathbf{x}),\quad \tilde{\mathbf{f}} = f((\mathbf{x}^\text{ft})^\text{deq}).
\end{equation}
To ensure that quantization does not introduce harmful feature drift, we minimize MSE between these feature representations:
\begin{equation}
\mathcal{L}_{\text{quant}} = \mathbb{E}_{\mathbf{x} \sim \mathcal{S}} \left[ \left\| \mathbf{f} - \tilde{\mathbf{f}} \right\|_2^2 \right],
\end{equation}
where \( \mathcal{S} \) denotes the set of synthetic images. The finetuned image \( \mathbf{x}^\text{ft} \) is optimized to minimize this loss, thereby aligning its quantized feature representation with that of the original.

In practice, we explore three refinement strategies in our framework: (1) applying refinement only before group quantization, (2) only after group quantization, and (3) both before and after. We analyze the quantitative effects of each strategy in our ablation study and provide qualitative examples in the final part of the experimental section through visualization.

\subsection{Storage Measurement}
\noindent
Under a fixed storage budget, our objective is to maximize the number of synthetic images per class. Starting from a baseline setting with IPC=\(m\), we apply our compression framework to reduce the per-class footprint to IPC=\(n\), thereby enabling a denser representation under the same storage constraint.

The total storage required to represent the synthetic data comprises three components:  
(i) \textbf{Group Indices} \( \mathcal{G} = \{ g_i \}_{i=1}^{P \times m} \), which indicate the group assignment for each patch,  
(ii) \textbf{Quantization Parameters} \( \mathcal{Q} = \{ \theta_g \}_{g=1}^{G} \), shared across patches within each group, and  
(iii) \textbf{Quantized Images} \( \mathcal{X}^q = \{ x^{\text{flat}}_g \}_{g=1}^G \), representing the flattened and quantized pixel values of each group.
To further reduce the footprint of \( \mathcal{X}^q \), we apply \textbf{Entropy Coding} (EC), which leverages the statistical redundancy in quantized values for additional compression.  

Given a target storage budget corresponding to IPC=\(n\), we ensure that the total compressed size does not exceed this constraint:
\begin{equation}
    size(\mathcal{G}) + size(\mathcal{Q}) + size(\text{EC}(\mathcal{X}^q)) \leq size(\text{IPC}).
\end{equation}

This formulation allows us to fit more quantized synthetic samples within the same memory budget, thereby improving the representational density of the condensed dataset.

\section{Experiment}
In this section, we evaluate the effectiveness of our proposed method in compressing synthetic images generated by various distillation techniques. We also conduct an ablation study to analyze the contribution of each component including group quantization, refinement, and entropy coding, to the overall performance.

\begin{table*}[ht]
\centering
\begin{minipage}{0.55\linewidth}
\centering
\footnotesize
\setlength{\tabcolsep}{3.5pt} 
\begin{tabular}{l|c|c|c|c|c|c}
\cmidrule[1pt]{1-7}
Dataset& I-Nette & I-Woof & I-Fruit & I-Meow &  I-Squawk & I-Yellow\\\cmidrule[0.5pt]{1-7}
MTT& \acc{63.0}{1.3}  & \acc{35.8}{1.8}  &\acc{40.3}{1.3} &\acc{40.4}{2.2}&\acc{52.3}{1.0}&\acc{60.0}{1.5}\\
DATM& \acc{65.8}{2.0} &  \acc{38.8}{1.6} &\acc{41.2}{1.4}  & \acc{45.7}{1.8} &\acc{56.3}{1.5} & \acc{61.1}{1.8}\\
HaBa & \acc{64.7}{1.6}&\acc{38.6}{1.3} & \acc{42.5}{1.5}&\acc{42.9}{0.9} & \acc{56.8}{1.0}&\acc{63.0}{1.6}\\
SPEED & \acc{72.9}{1.5}& \acc{44.1}{1.4}&\acc{50.0}{0.8} &\acc{52.0}{1.3}&\acc{71.8}{1.3}&\acc{70.5}{1.5}\\
FreD  & \acc{72.0}{0.8} &\acc{41.3}{1.2}&\acc{47.0}{1.1}&\acc{48.6}{0.4}&\acc{67.3}{0.8}&\acc{69.2}{0.6}\\
AutoPalette&\acc{73.2}{0.6}&\acc{44.3}{0.9}&\acc{48.4}{1.8}&\acc{53.6}{0.7}&\acc{68.0}{1.4}&\acc{72.0}{1.6}\\\cmidrule[0.5pt]{1-7}
 DATM+Ours & \colorLC \textbf{\acc{81.1}{1.2}}&\colorLC\textbf{\acc{53.0}{0.8}}&\colorLC\textbf{\acc{56.6}{0.9}}&\colorLC\textbf{\acc{61.2}{0.9}}&\colorLC\textbf{\acc{80.6}{0.7 }}&\colorLC\textbf{\acc{78.9}{0.9}}\\
\cmidrule[1.pt]{1-7}
\end{tabular}
\caption{Performance comparison on ImageNet Subsets (I-\textit{subset}).}
\label{isubset}

\end{minipage}%
\hfill
\begin{minipage}{0.45\linewidth}
\centering
\footnotesize
\setlength{\tabcolsep}{3.5pt} 
\begin{tabular}{l|cc|cc}
\cmidrule[1.0pt]{1-5}
Dataset & \multicolumn{2}{c|}{CC3M} &\multicolumn{2}{c}{Places365 Standard} \\\cmidrule[0.5pt]{1-5}
IPC &  1 & 10 & 1& 10\\\cmidrule[0.5pt]{1-5}
 Random& \acc{2.9}{0.1}& \acc{7.0}{0.1} & \acc{0.9}{0.1}& \acc{3.9}{0.1}  \\
DM & \acc{5.9}{0.2} & \acc{8.7}{0.1}&\acc{2.6}{0.1} & \acc{5.8}{0.1} \\
IDC & \acc{8.2}{0.3} & \acc{12.1}{0.2}&\acc{4.0}{0.1}& \acc{9.8}{0.1}\\\cmidrule[0.5pt]{1-5}
DM+Ours& \colorLC \textbf{\acc{9.9}{0.2}} & \colorLC\textbf{\acc{17.4}{0.1}}  & \colorLC \textbf{\acc{7.7}{0.1}} & \colorLC\textbf{\acc{19.2}{0.1}} \\\cmidrule[0.5pt]{1-5}
Whole $\mathcal{T}$ & \multicolumn{2}{c|}{\acc{46.7}{0.1}} & \multicolumn{2}{c}{\acc{40.4}{0.1}}\\\cmidrule[1.0pt]{1-5}
\end{tabular}
\caption{Performance comparison on real-world datasets.}
\label{cc3m} 
\end{minipage}
\end{table*}

\begin{table}[]
\centering
\begin{minipage}{0.4\linewidth}
\centering
\footnotesize
\setlength{\tabcolsep}{3.5pt} 
\begin{tabular}{l|c|c}\cmidrule[1pt]{1-3}
Method & IPC=10 & IPC=20\\\cmidrule[0.5pt]{1-3}
Random & 42.6 & 57.0 \\
Herding &56.2 & 72.9\\
DM & 69.1 & 77.2 \\
IDC & 82.9 & 86.6 \\
DDiF & 90.5 & 92.7 \\
 DM+Ours & \colorLC \textbf{93.1} & \colorLC \textbf{95.0} \\\cmidrule[0.5pt]{1-3}
 Whole $\mathcal{T}$&\multicolumn{2}{c}{93.4}
\\\cmidrule[1pt]{1-3}
\end{tabular}
\caption{Performance comparison on Audio domain.}
\label{audio}
\end{minipage}%
\hfill
\begin{minipage}{0.5\linewidth}
\centering
\footnotesize
\setlength{\tabcolsep}{3pt} 
\rule{0pt}{2.ex}
\begin{tabular}{c|l|c|c}
\cmidrule[1pt]{1-4}
$\mathcal{L}$ & Method & MNet & SNet\\\cmidrule[0.5pt]{1-4}
\multirow{3}{*}{DC} & IDC & 78.7&79.9   \\
&  DDiF & 87.1  & 89.6\\
&  Ours & \colorLC \textbf{93.7}& \colorLC \textbf{91.1}\\\cmidrule[0.5pt]{1-4}
\multirow{3}{*}{DM} & IDC &85.6&85.3  \\
&  DDiF &88.4&93.1 \\
& Ours& \colorLC  \textbf{93.9} & \colorLC \textbf{93.1}\\\cmidrule[0.5pt]{1-4}
 \multicolumn{2}{c|}{Whole $\mathcal{T}$}&91.6& 98.3
\\\cmidrule[1.0pt]{1-4}
\end{tabular}
\caption{Performance comparison on 3D voxel domain at IPC=1.}
\label{3d}
\end{minipage}
\end{table}

\subsection{Experimental Setting}
We evaluate our method across diverse datasets, ranging from toy datasets like CIFAR~\cite{cifar} to large-scale, real-world datasets such as ImageNet subsets~\mtt, CC3M~\cite{cc3m}, and Places365~\cite{place365}. As a plug-and-play framework, our method is applied to synthetic images generated by various distillation approaches, including DM~\dm, DSA~\dsa, and DATM~\datm. Given a storage budget of IPC=\(m\), we first generate synthetic images using existing methods, then compress them to IPC=\(n\) using our framework.

By default, we apply 2-bit quantization using non-overlapping patches of size \(5 \times 5\). We perform a grid search to determine the maximum number of groups that satisfy the storage constraint. Prior to group quantization, synthetic images are refined for 500 iterations on CIFAR-10/100 and 2000 iterations on ImageNet subsets. For evaluation, models are trained on the compressed synthetic datasets and tested on the original test sets.

\subsection{Experimental Results}
\paragraph{Results on small-scale datasets.}  
We compare our quantization-based parameterization method against various PDC baselines, including Factorization~\cite{liu2022dataset} (HaBa), Coding Matrices~\cite{wei2023sparse} (SPEED), Frequency Domain~\cite{shin2023frequency} (FreD), Spectral Domain~\cite{neuralspectral} (Spectral), and Color Redundancy Reduction~\cite{yuan2024colororiented} (AutoPalette). 

As shown in \reftab{cifarexp}, our framework significantly improves the performance of DM and DSA. For instance, at IPC=1, DM achieves 26\% accuracy, whereas DM+Ours achieves 54.1\%, a relative gain of over 100\%. A similar trend is observed for DSA. When applied to the latest distillation method DATM, our framework achieves state-of-the-art performance across all IPC levels.
Although AutoPalette~\cite{yuan2024colororiented} reduces color redundancy via an additional network, it incurs higher computational cost. Despite sharing the same distillation base (DATM), our method consistently outperforms AutoPalette, demonstrating both higher efficiency and accuracy.

\paragraph{Results on mid- and large-scale datasets.}  
Following~\mtt, we evaluate our method on high-resolution ImageNet subsets. As shown in \reftab{isubset}, integrating our framework into DATM consistently yields the best performance across all subsets, with gains ranging from 6.9\% (I-Yellow) to 12.6\% (I-Squawk). Furthermore, when evaluating our method on CC3M and Places365, we observe the highest test accuracy in all cases (see \reftab{cc3m}). These results demonstrate the scalability of our framework to more complex domains.

\paragraph{Generalization to other modalities.}  
To demonstrate versatility, we apply our framework to audio and 3D data domains following the experiment setup in \cite{ddif}. As shown in \reftab{audio} and \reftab{3d}, our method outperforms recent approaches such as DDiF~\cite{ddif}, highlighting its cross-modal robustness.

\begin{table}[]
\centering
\footnotesize
\setlength{\tabcolsep}{3.5pt} 
\begin{tabular}{l|cccc}\cmidrule[1pt]{1-5}
\multirow{2}{*}{Method}  & \multicolumn{4}{c}{Network}  \\
&   ConvNet & AlexNet & VGG11 & ResNet18 \\\cmidrule[0.5pt]{1-5}

DC & \acc{44.9}{0.5}  &\acc{22.4}{1.4}& \acc{35.9}{0.7} & \acc{18.4}{0.4} \\
MTT  & \acc{65.3}{0.7} & \acc{34.2}{0.6} & \acc{50.3}{0.8} &  \acc{46.6}{0.6}\\
DATM & \acc{66.8}{0.2} & \acc{32.7}{3.8} &\acc{38.8}{2.1} & \acc{51.4}{1.7}\\
FRePo  & \acc{65.5}{0.4} & \acc{61.9}{0.7} & \acc{59.4}{0.7}& \acc{58.1}{0.6}\\
Spectral &\acc{73.4}{0.2} & \acc{71.4}{0.3}  & \acc{67.8}{0.2} & \acc{64.9}{1.3}\\\cmidrule[0.5pt]{1-5}
DATM+Ours & \colorLC\textbf{\acc{79.0}{0.3}}&\colorLC\textbf{\acc{74.7}{0.5}} &\colorLC\textbf{\acc{74.7}{0.3}} &\colorLC\textbf{\acc{77.5}{0.3}} \\\cmidrule[1pt]{1-5}
\end{tabular}
\caption{Cross-architecture performance comparison between our framework and other methods at IPC=10.}

\label{ablationcross}
\end{table}

\paragraph{Cross-architecture validation.}  
We further assess generalization across architectures. Using synthetic images generated by DATM, we apply our method and evaluate on AlexNet~\cite{alexnet}, VGG~\cite{vgg}, and ResNet~\cite{resnet}. As shown in \reftab{ablationcross}, our method demonstrates superior generalization performance compared to previous approaches.

\begin{table}[h]
\footnotesize
\centering
\begin{tabular}{c|c|c|c|c}\cmidrule[1pt]{1-5} 
GAQ & Refinement & EC &  CIFAR-10 & I-Nette\\\cmidrule[0.5pt]{1-5} 
\xmark& \xmark & \xmark &\acc{71.8}{0.3} &\acc{75.2}{0.8} \\
\cmark& \xmark & \xmark &\acc{76.1}{0.3}& \acc{76.5}{1.1}\\
\cmark& \cmark & \xmark&\acc{77.2}{0.3}&\acc{77.2}{1.0} \\
\cmark& \cmark & \cmark &\colorLC\textbf{\acc{79.0}{0.3}}&\colorLC\textbf{\acc{81.1}{1.2}} \\
\cmidrule[1pt]{1-5} 
\end{tabular}
\caption{
Effect of each component on final performance. The first row shows the baseline using global asymmetric quantization. IPC is set to 10 for both CIFAR-10 and I-Nette.
}

\label{ablation_eff}
\end{table}
\begin{table}[h]
\centering
\footnotesize
\begin{tabular}{l|c|c|c}\cmidrule[1pt]{1-4} 
Refinement before GAQ & \cmark  & \xmark & \cmark \\
Refinement after GAQ & \xmark & \cmark &\cmark\\ \cmidrule[0.5pt]{1-4} 
Test Accuracy (\%) & \acc{68.9}{0.4} & \acc{68.7}{0.5} & \acc{68.9}{0.3}\\
\cmidrule[1pt]{1-4} 
\end{tabular}
\caption{Test accuracy on CIFAR-10 at IPC=1 with refinement applied at different stages.}

\label{whereto}
\end{table}
\begin{figure}[h]
\centering
\includegraphics[height=2in]{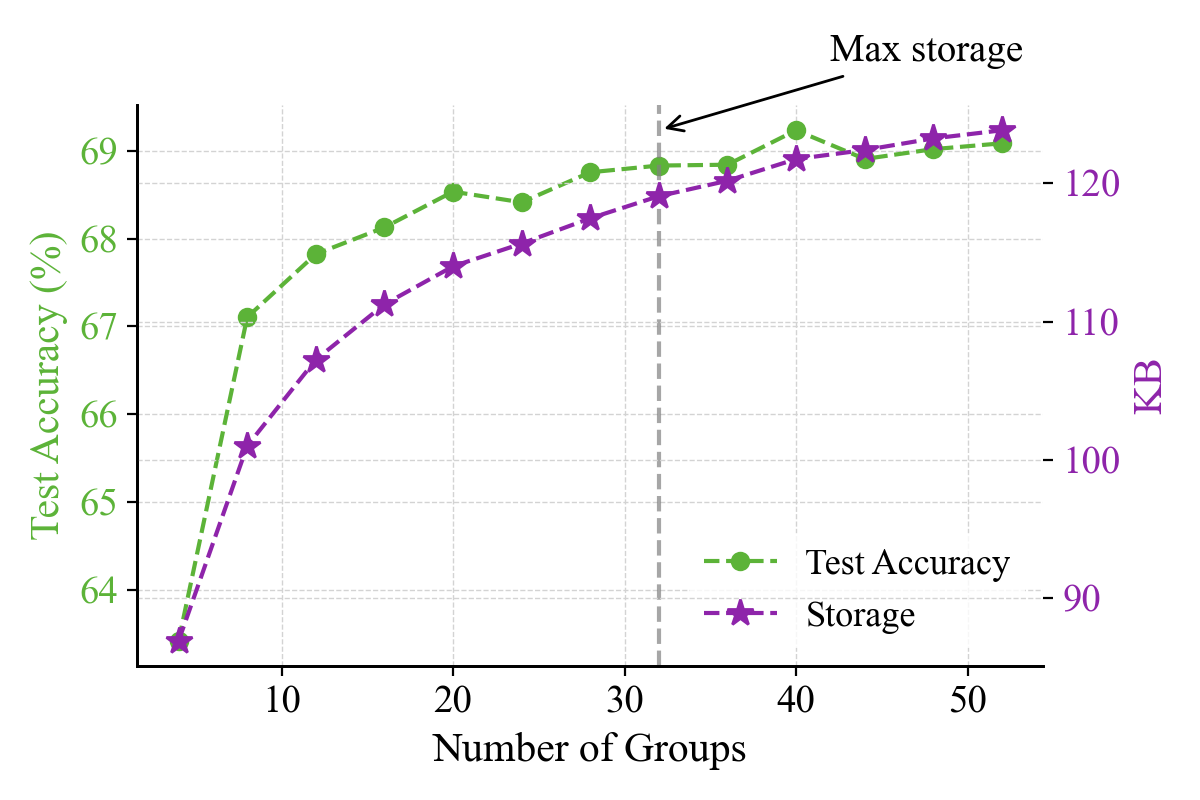}
\caption{
Measured test accuracy (\%) and storage across different numbers of groups. The vertical dotted line marks the maximum allowable storage under the given budget.
}

\label{storagesize}
\end{figure}

\subsection{Ablation Study}
We conduct an ablation study to analyze the effect of each component in our framework on the final performance, as well as the impact of the number of groups. By default, we adopt DATM to generate synthetic images.

\paragraph{Effect of Each Component.}  
As shown in \reftab{ablation_eff}, applying asymmetric quantization (AQ) alone achieves 71.8\% test accuracy on CIFAR-10, outperforming the base DATM method (66.8\%) at IPC=10. Adding group-aware quantization (GAQ) further improves accuracy to 76.1\%, and incorporating the refinement module increases it to 77.2\%. Finally, applying entropy coding allows more samples to be stored within the same budget, leading to a final performance of 79\%.

\paragraph{Ablation on Refinement Timing.}  
As shown in \reftab{whereto}, applying refinement before group quantization yields better performance than applying it afterward. This is likely because group quantization assigns patches to groups based on their quantization parameters, which are more accurately estimated when the image is refined beforehand. In contrast, applying group quantization first may lead to suboptimal grouping due to uncorrected quantization noise. Notably, performing refinement both before and after group quantization does not lead to further improvement, suggesting diminishing returns from post-group refinement.

\paragraph{Ablation on the Number of Groups.}  
\reffig{storagesize} shows how the number of groups ($G$ in Eq. (6)) affects performance. We observe that increasing the number of groups consistently improves test accuracy, while staying within the storage budget. Based on this trend, we choose the maximum number of groups that satisfies the storage constraint.

\begin{figure}[]
\centering
\includegraphics[height=2.5in]{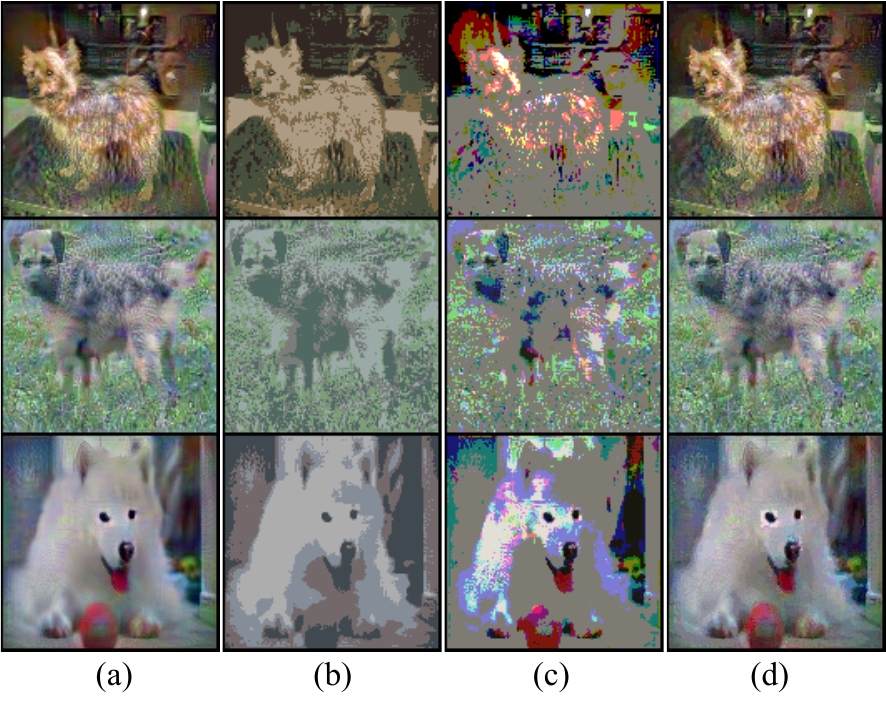}
\caption{
Visualization of (a) original synthetic images and their quantized versions using (b) Median Cut, (c) Asymmetric Quantization (AQ), and (d) Group-Aware Quantization (GAQ). While Median Cut preserves texture but distorts color, and AQ preserves color but loses fine details, GAQ achieves a better balance between texture and color fidelity.
}
\label{vis}
\end{figure}

\paragraph{Visualization of Synthetic Images.}  
We visualize quantized synthetic images generated by MTT~\cite{cazenavette2022distillation} on I-Woof. As shown in \reffig{vis}, the Median Cut method retains texture but alters color severely, while AQ preserves color at the cost of heavy texture distortion. In contrast, our group-based quantization strikes a better balance, maintaining both texture and color fidelity. Additional examples are provided in the supplementary material.

\section{Conclusion}
We propose a novel quantization framework for dataset condensation that achieves substantial storage reduction while preserving downstream performance. Unlike prior work that focuses solely on synthetic image generation, our method explicitly targets bit-level redundancy through patch-based quantization, group-wise parameter sharing, and lightweight feature refinement. This combination enables effective compression even at extremely low bit-widths. Extensive experiments across diverse distillation methods and datasets validate that our approach consistently outperforms existing parameterized condensation baselines under tight storage constraints.

\paragraph{Limitation and Future Work.}
Our framework is currently designed for condensation pipelines that focus on visual data and hard labels. It does not yet support methods that utilize soft labels. Future work may explore label-aware quantization. 
In addition, our method is tailored for CNN-based architectures operating on spatial patches, and has not been extended to transformers. 
Designing quantization strategies that operate on token embeddings presents a promising direction for transformer-based models.

\section{Acknowledgments}
This work was supported by the National Research Foundation of Korea (NRF) grant funded by the Korea government (MSIT) (No. RS-2025-24534076) and (No. RS-2025-02216217), and by the IITP (Institute of Information \& Communications Technology Planning \& Evaluation) - ITRC (Information Technology Research Center) grant funded by the Korea government (Ministry of Science and ICT) (IITP-2025-RS-2024-00438239)

\bibliography{aaai2026}
\end{document}